\documentclass[9pt,twocolumn,twoside]{pnas-altered-arxiv}

\templatetype{pnasresearcharticle} 
\usepackage{dsfont}
\usepackage{soul}
\usepackage{amsmath}
\usepackage{appendix}

\newcommand{\sref}[1]{Sec.~\ref{#1}}

\newcommand{\kn}[1]{{\textcolor{black}{#1}}}
\newcommand{\kkn}[1]{{\textcolor{black}{#1}}}

\title{Veridical Data Science}
\author[a,b,c]{Bin Yu}
\author[a]{Karl Kumbier}
\affil[a]{Statistics Department, University of California, Berkeley, CA 94720}
\affil[b]{EECS Department, University of California, Berkeley, CA 94720}
\affil[c]{Chan Zuckerberg Biohub, San Francisco, CA 94158}
\leadauthor{Yu} 

\begin{abstract}
Building and expanding on principles of statistics, machine learning, and scientific inquiry, we propose the predictability, computability, and stability (PCS) framework for veridical data science. Our framework, comprised of both a workflow and documentation, aims to provide responsible, reliable, reproducible, and transparent results across the entire data science life cycle. The PCS workflow uses predictability as a reality check and considers the importance of computation in data collection/storage and algorithm design. It augments predictability and computability with an overarching stability principle for the data science life cycle. Stability expands on statistical uncertainty considerations to assess how human judgment calls impact data results through data and model/algorithm perturbations. Moreover, we develop inference procedures that build on PCS, namely PCS perturbation intervals and PCS hypothesis testing, to investigate the stability of data results relative to problem formulation, data cleaning, modeling decisions, and interpretations. We illustrate PCS inference through neuroscience and genomics projects of our own and others and compare it to existing methods in high dimensional, sparse linear model simulations. Over a wide range of misspecified simulation models, PCS inference demonstrates favorable performance in terms of ROC curves. Finally, we propose PCS documentation based on R Markdown or Jupyter Notebook, with publicly available, reproducible codes and narratives to back up human choices made throughout an analysis. The PCS workflow and documentation are demonstrated in a genomics case study available on Zenodo \cite{yu_kumbier_2018}.
\end{abstract}

\dates{This manuscript was compiled on \today}

\begin{document}
\maketitle
\thispagestyle{firststyle}
\abscontentformatted

\section{Introduction}

Data science is a field of evidence seeking that combines data with domain
information to generate new knowledge. The data science life cycle (DSLC) begins with a domain question or problem and proceeds through collecting, managing,
processing (cleaning), exploring, modeling, and interpreting\footnote{For a precise definition of interpretability in the context of machine learning, we refer to our recent paper \cite{murdoch2019interpretable}} data results to guide new
actions (Fig. \ref{fig:lifecycle}). Given the trans-disciplinary nature of this
process, data science requires human involvement from those who collectively understand both
the domain and tools used to collect, process, and model data. These
individuals make implicit and explicit judgment calls throughout the DSLC. The limited transparency in reporting such judgment calls has blurred the evidence for many analyses, resulting in more false-discoveries than might otherwise occur \cite{stark2018cargo,ioannidis2005most}. This fundamental issue necessitates veridical data science to extract reliable and reproducible information from data, with an enriched technical language to communicate and evaluate empirical evidence in the context of human decisions. Three core principles: predictability, computability, and stability (PCS) provide the foundation for such a data-driven language and a unified data analysis framework. They serve as minimum requirements for veridical data science\footnote{Veridical data science is the broad aim of our proposed framework (veridical meaning ``truthful'' or ``coinciding with reality''). This paper has been on arXiv since Jan. 2019 under the old title ``Three principles of data science: predictability, computability, stability (PCS).}.

Many ideas embedded in PCS have been widely used across various areas of data science. Predictability plays a central role in science through
Popperian falsifiability \cite{popperp1959logic}. If a model does not accurately predict new observations, it can be rejected or updated. Predictability has been adopted by the machine learning community as a goal of its own right
and more generally to evaluate the quality of a model or data result
\cite{breiman2001statistical}. While statistics has always considered prediction, machine learning emphasized its importance for empirical rigor. This was in large part powered by computational advances that made it possible to compare models
through cross-validation (CV), developed by statisticians Stone and
Allen \cite{stone1974cross, allen1974relationship}.

The role of computation extends beyond prediction, setting limitations on how
data can be collected, stored, and analyzed. Computability has played an
integral role in computer science tracing back to Alan Turing's seminal work on
the computability of sequences \cite{turing1937computable}. Analyses of
computational complexity have since been used to evaluate the tractability of
machine learning algorithms \cite{hartmanis1965computational}.
Kolmogorov built on Turing's work through the notion of Kolmogorov complexity,
which describes the minimum computational resources required to represent an
object \cite{li2008introduction, kolmogorov1963tables}. Since Turing machine-based 
notions of computabiltiy are not computable in practice, we
treat computability as an issue of algorithm efficiency and scalability. 
This narrow definition of computability addresses computational considerations at the modeling stage of the DSLC but does not deal with data collection, storage, or cleaning.

Stability\footnote{We differentiate
between the notions of stability and robustness as used in statistics. The
latter has traditionally been used to investigate performance of statistical
methods across a range of distributions, while the former captures a much
broader range of perturbations throughout the DSLC as
discussed in this paper. At a high level, stability is about robustness.} is a common sense principle and a prerequisite for knowledge. It is related to the
notion of scientific reproducibility, which Fisher and Popper argued is a
necessary condition for establishing scientific results \cite{popperp1959logic,
fisher1937design}. While replicability across laboratories has long been an
important consideration in science, computational reproducibility has come to
play an important role in data science as well. For example,
\cite{donoho2009reproducible} discusses reproducible research in the context of
computational harmonic analysis.  More broadly, \cite{stark2018before} advocates
for ``preproducibility'' to explicitly detail all steps along the DSLC and ensure sufficient information for quality control. Stability at the modeling stage of the DSLC has been advocated in \cite{yu2013stability} as a minimum requirement for
reproducibility and interpretability. Modeling stage stability 
unifies numerous previous works, including Jackknife, subsampling, bootstrap sampling, robust statistics, semi-parametric statistics, and Bayesian sensitivity analysis (see
\cite{yu2013stability} and references therein). These methods have been enabled in practice through computational advances and allow researchers to investigate the reproducibility of data results. Econometric models with partial identification (see the book \cite{manski2013public} and references therein) and fundamental theoretical results in statistics, such as the central limit theorem (CLT), can also be viewed as stability considerations.

In this paper, we unify and expand on these ideas through the PCS framework, which is built on the three principles of data science. The PCS framework consists of PCS workflow and transparent PCS documentation. It uses predictability as a reality check, computability to ensure that the DSLC is tractable, and stability to test the reproducibility of data results (\sref{sec:pcs}) relative to human judgment calls at every step of the DSLC. In particular, we develop basic PCS inference, which leverages data and model perturbations to evaluate the uncertainty human decisions introduce into the DSLC  (\sref{sec:inference}). We propose PCS documentation in R MarkDown or a Jupyter (iPython) Notebook to justify these decisions through narratives, code and visualizations (\sref{sec:doc}). We draw connections between causal inference and the PCS framework, demonstrating the utility of the latter as a recommendation system for generating scientific hypotheses (\sref{sec:hypothesis}). We conclude by discussing areas for further work, including additional vetting of the framework and theoretical analyses on connections between the three principles. A case study of our proposed framework based on the authors' work studying gene regulation in \textit{Drosophila} is documented on \href{https://doi.org/10.5281/zenodo.3522419}{Zenodo}.
\section{PCS principles in the DSLC}\label{sec:pcs}
\begin{figure}
  \includegraphics[width=0.49\textwidth]{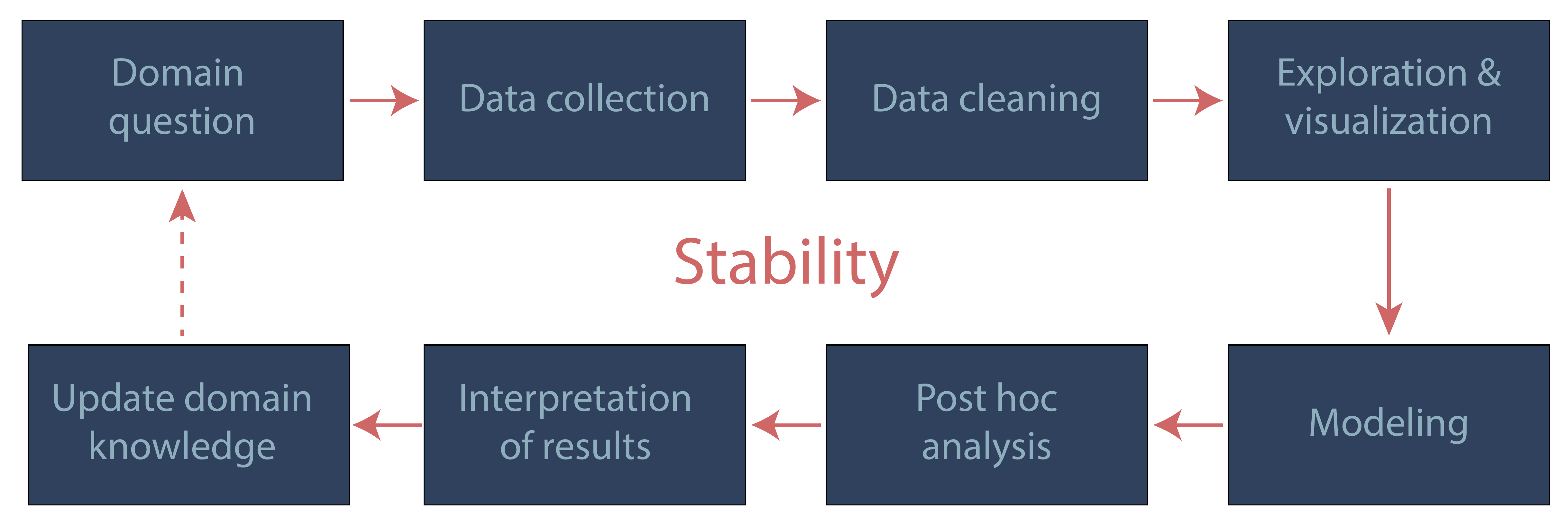}
  \caption{The data science life cycle}
  \label{fig:lifecycle}
\end{figure}

Given a domain problem and data, the purpose of the DSLC is to generate
knowledge, conclusions, and actions (Fig. \ref{fig:lifecycle}). The PCS framework aims to ensure that this process is both reliable and reproducible through the three fundamental principles of data science.  
Below we discuss the roles of the three principles within the PCS framework\footnote{We organize our discussion with respect to the steps in the DSLC.}, including PCS workflow and PCS documentation. The former applies the relevant principles at every step of the DSLC, with stability as the paramount consideration, and contains PCS inference proposed in \sref{sec:inference}. The latter documents the PCS workflow and judgment calls made with a 6-step format described in \sref{sec:doc}.

\subsection{Stability assumptions initiate the DSLC}
The ultimate goal of the DSLC is to generate knowledge that
is useful for future actions, be it a biological
experiment, business decision, or government policy. Stability is a useful
concept to address whether another researcher making alternative, appropriate\footnote{We use the term appropriate to mean well-justified from domain knowledge and an understanding of the data generating process. The term ``reasonable'' has also been used with this definition \cite{yu2013stability}.} decisions would obtain similar conclusions. At the modeling stage, stability has previously been advocated in \cite{yu2013stability}.  In this context, stability refers to acceptable consistency of a data result relative to appropriate perturbations
of the data or model. For example, jackknife
\cite{quenouille1949problems,quenouille1956notes,tukey1958bias}, bootstrap
\cite{efron1992bootstrap}, and cross validation
\cite{stone1974cross,allen1974relationship} may be considered appropriate
perturbations if the data are deemed approximately independent and identically
distributed (i.i.d.) based on domain knowledge and an understanding of the data
collection process. 

Human judgment calls prior to modeling also impact data results. The validity of an analysis relies on implicit stability assumptions that allow data to be treated as an informative representation of some natural phenomena.  When these assumptions do not hold in a particular domain, conclusions rarely generalize to new settings unless empirically proven by future data. This makes it essential to evaluate stability to guard against costly future actions and false discoveries, particularly in the domains of science, business, and public policy, where data results are used to guide large scale actions, and in medicine, where human lives are at stake. Below we outline stability considerations that impact the DSLC prior to modeling.

\textbf{Question or problem formulation:} The DSLC begins with a domain problem or a question, which could be hypothesis-driven or discovery-based. For instance, a biologist may want to discover biomolecules that regulate a gene's expression. In the DSLC this question must be translated into a question regarding the output of a model or analysis of data that can be measured/collected. There are often multiple translations of a domain problem into a data science problem. For example, the biologist described above could measure factors binding regulatory regions of the DNA that are associated with the gene of interest. Alternatively, she could study how the gene covaries with regulatory factors across time and space. From a modeling perspective, the biologist could identify important features in a random forest or through logistic regression. Stability relative to question or problem formulation implies that the domain conclusions are qualitatively consistent across these different translations.

\textbf{Data collection:} To answer a domain question, domain experts and data scientists collect data based on prior knowledge and available resources. When this data is used to guide future decisions, researchers implicitly assume that the data is relevant to a future time. In other words, they assume that conditions affecting data collection are stable, at least relative to some aspects of the data. For instance, if multiple laboratories collect data to answer a domain question, protocols must be comparable across experiments and laboratories if they expect to obtain consistent results. These stability considerations are closely related to external validity in medical research, which characterizes similarities between subjects in a study and subjects that researchers hope to generalize results to.  We will discuss this idea more in \sref{sec:pred}.

\textbf{Data cleaning and preprocessing:} Statistics and machine learning models or algorithms help data scientists answer domain questions. 
Using models or algorithms requires cleaning
(pre-processing) raw data into a suitable format, be it a categorical
demographic feature or continuous measurements of biomarker concentrations. For
instance, when data come from multiple laboratories, biologists must decide
how to normalize individual measurements (for example see
\cite{bolstad2003comparison}).  When data scientists preprocess data,
they are implicitly assuming that their choices are not unintentionally biasing the essential information in the raw data. In other words, they assume
that the knowledge derived from a data result is stable with respect to their
processing choices. If such an assumption cannot be justified, they should use
multiple appropriate processing methods and interpret data results that are
stable across these methods.  Others have advocated evaluating results across
alternatively processed datasets under the name ``multiverse analysis''
\cite{steegen2016increasing}. Although the stability principle was developed
independently of this work, it naturally leads to a multiverse-style analysis.

\textbf{Exploratory data analysis:} Both before the modeling stage and in 
post hoc analyses, data scientists often engage in exploratory data analysis (EDA) to identify interesting relationships in the data and interpret data results. When
visualizations or summaries are used to communicate these analyses, it is
implicitly assumed that the relationships or data results are stable with
respect to any decisions made by the data scientist.  For example, if the
biologist believes that clusters in a heatmap represent biologically meaningful
groups, she should expect to observe the same clusters with respect to any
appropriate choice of distance metric, data perturbation, or clustering method.

\subsection{Predictability as reality check}\footnote{Predictability is a form of empirical validation, though other reality checks may be performed beyond prediction (e.g. checking whether a model recovers known phenomena).}\label{sec:pred}
After data collection, cleaning/preprocessing, and EDA, models or
algorithms\footnote{Different model or algorithm choices could correspond to different
translations of a domain problem.} are frequently used to identify more complex
relationships in data.  Many essential components of the modeling stage rely on
the language of mathematics, both in technical papers and in code.  A seemingly
obvious but often ignored question is why conclusions presented in the language
of mathematics depict reality that exists independently in nature, and to what
extent we should trust mathematical conclusions to impact this external
reality.\footnote{The PCS documentation in \sref{sec:doc} helps users
assess whether this connection is reliable.} 

This concern has been articulated and addressed
by many others in terms of prediction. For instance, Philip Dawid drew connections between statistical inference and prediction under the name ``prequential statistics,'' highlighting the importance of forecasts in statistical analyses \cite{dawid1984present}. David Freedman argued that when a model's predictions are not tested against reality, conclusions drawn from the model are unreliable \cite{freedman1991statistical}. Seymour Geisser advocated that
statistical analyses should focus on prediction rather than parametric
inference, particularly in cases where the statistical model is an inappropriate
description of reality \cite{geisser1993predictive}. Leo Breiman
championed the essential role of prediction in developing realistic models that
yield sound scientific conclusions \cite{breiman2001statistical}. It can even be
argued that the goal of most domain problems is prediction at the meta level. That is, the primary value of learning relationships in data is often to predict some
aspect of future reality.

\subsubsection{Formulating prediction} 
We describe a general framework for prediction with data $D=(\mathbf{x}, y)$,
where $\mathbf{x}\in\mathcal{X}$ represents input features and $y\in\mathcal{Y}$
the prediction target. Prediction targets $y\in \mathcal{Y}$ may be observed
responses (e.g. supervised learning) or extracted from data (e.g. unsupervised
learning). Predictive accuracy is a simple, quantitative metric to
evaluate how well a model represents relationships in $D$. It is well-defined
relative to a prediction function, testing data, and an evaluation metric. We
detail each of these elements below.

\vspace{0.5em}
\noindent\textbf{Prediction function:} The prediction function
\begin{equation} 
h: \mathcal{X} \rightarrow \mathcal{Y}
\label{eq:prediction}
\end{equation} 
represents relationships between the observed features and the prediction
target. For instance, in the case of supervised learning $h$ may be a linear
predictor or decision tree. In this setting, $y$ is typically an observed
response, such as a class label. In the case of unsupervised learning, $h$ could
map from input features to cluster centroids. 

To compare multiple prediction functions, we consider

\begin{equation}
  \{h^{(\lambda)}: \lambda \in \Lambda\},
\label{eq:predfuns}
\end{equation}
where $\Lambda$ denotes a collection models/algorithms. For example, $\Lambda$
may define different tuning parameters in lasso \cite{tibshirani1996regression}
or random forest \cite{breiman2001random}. For deep neural networks, $\Lambda$ could describe different network architectures. For algorithms with a randomized
component, such as k-means or stochastic gradient descent, $\Lambda$ can
represent repeated runs.  More broadly, $\Lambda$ may describe a set of competing algorithms such as linear models, random forests, and neural networks, each corresponding to a different problem translations. We discuss model
perturbations in more detail in \sref{sec:model}.

\vspace{0.5em}
\noindent\textbf{Testing (held-out) data:} We distinguish between
\textit{training data} that are used to fit a collection of prediction
functions, and \textit{testing data} that are used to evaluate the
accuracy of fitted prediction functions.\footnote{\kn{In some settings, a third 
set of data are used to tune model parameters.}} At a minimum, one should evaluate predictive accuracy on a held-out test set generated at the same time and under the same conditions as the training data (e.g. by randomly sampling a subset of observations). This type of assessment addresses questions internal validity, which describe the strength of a relationship in a given sample. It is also often important to understand how a model will perform in future conditions that differ from those that generated the training data. For instance, a biologist may want to apply their model to new cell lines. A social scientist might use a model trained on residents from one city to predict the behavior of residents in another. As an extreme example, one may want to use transfer learning to apply part of their model to an entirely new prediction problem. Testing data gathered under different conditions from the training data directly addresses questions of external validity, which describe how well a result will generalize to future observations.  Domain knowledge and/or empirical validation are essential to assess the appropriateness of different prediction settings. These decisions should  be reported in the proposed PCS documentation (\sref{sec:doc}).

\vspace{0.5em}
\noindent\textbf{Prediction evaluation metric:} The prediction evaluation metric
\begin{equation} 
\ell:\mathcal{H}\times \mathcal{X} \times
\mathcal{Y}\rightarrow \mathbb{R}_{+} 
\label{eq:loss} 
\end{equation} 
quantifies the accuracy of a prediction function $h \in\mathcal{H}$ by measuring
the similarity between $h(\mathbf{x})$ and $y$. We adopt the convention that increasing values of $\ell(h,\mathbf{x},y)$ imply worse predictive accuracy. The prediction evaluation metric should be selected to reflect
domain-specific considerations, such as the types of errors that are more costly. \kkn{In fact, there is an entire area of research devoted to evaluating the quality of probabilistic forecasts through ``scoring rules'' (see \cite{gneiting2007strictly} and references therein).} 
In some cases, it may be appropriate to consider
multiple prediction evaluation metrics and focus on models that are deemed
accurate with respect to all. 

\vspace{0.5em}
Prediction requires human input to
formulate, including the preferred structure of a model/algorithm and what it means for a model to be suitably accurate. For example, the biologist studying gene regulation may believe that the simple rules learned by decision trees are an an appealing
representation of interactions that exhibit thresholding behavior
\cite{wolpert1969positional}. If she is interested in a particular
cell-type, she may evaluate prediction accuracy on
test data measuring only these environments. If her responses are
binary with a large proportion of class-$0$ responses, she may choose an
evaluation function $\ell$ to handle the class imbalance. All of these decisions should be documented and argued for (\sref{sec:doc}) so that other researchers can review and assess the strength of conclusions based on transparent evidence.  The accompanying PCS
\href{https://doi.org/10.5281/zenodo.3522419}{documentation} provides a
detailed example.

\subsubsection{Cross validation}

As alluded to earlier, CV has become a powerful work horse to
select regularization parameters when data are approximately i.i.d. \cite{stone1974cross, allen1974relationship}.
CV divides data into blocks of observations, trains a model on all but one
block, and evaluates the prediction error over each held-out block. In other words, CV
incorporates the scientific principle of replication by evaluating whether a model accurately predicts the responses of
pseudo-replicates.  CV works more effectively as a tool to select regularization
parameters than as an estimate of prediction error, where it can incur high
variability due to the often positive dependencies between the estimated
prediction errors in the summation of the CV error \cite{fushiki2011estimation}.
Just as peer reviewers make judgment calls on whether a lab's experimental
conditions are suitable to replicate scientific results, data scientists must
determine whether a removed block represents a justifiable pseudo replicate of
the data, which requires information from the data collection process and domain
knowledge.

\subsection{Computability} 
In a broad sense, computability is the gate-keeper of data science. If data
cannot be generated, stored, managed, and analyzed efficiently and scalably,
there is no data science. Modern science relies heavily on information
technology as part of the DSLC.  Each step, from raw data
collection and cleaning, to model building and evaluation, rely on computing
technology and fall under computability in a broad sense. In a narrow sense,
computability refers to the computational feasibility of algorithms or model
building.

Here we use computability in the narrow-sense, which is closely associated with the rise of machine learning over the last three decades. Just as scientific
instruments and technologies determine what processes can be effectively
measured, computing resources and technologies determine the types of analyses
that can be carried out. In particular, computability is necessary to carry out predictability and stability analyses within the PCS framework. Computational constraints can also serve as a device for regularization. For example, stochastic gradient descent is widely used for optimization in machine learning problems \cite{robbins1951stochastic}. Both the stochasticity and early stopping of a stochastic gradient algorithm play the role of implicit regularization.

Computational considerations and algorithmic analyses have long been an
important part of statistics and machine learning. Even before digital
computing, calculus played a computational role in statistics through Taylor
expansions applied to different models. In machine learning, computational
analyses consider the number of operations and required storage space in terms
of observations $n$, features $p$, and tuning (hyper) parameters. When the
computational cost of addressing a domain problem or question exceeds available
computational resources, a result is not computable.  For instance, the
biologist interested in gene regulation may want to model interaction effects in a
supervised learning setting. However, there are $O(p^s)$ possible order-$s$
interactions among $p$ regulatory factors.  For even a moderate number of
factors, exhaustively searching for high-order interactions is not computable.
In such settings, data scientists must restrict modeling decisions to draw
conclusions. Thus it is important to document why certain restrictions were
deemed appropriate and the impact they may have on conclusions (\sref{sec:doc}).

Increases in computing power also provide an unprecedented opportunity to
enhance analytical insights into complex natural phenomena. We can now store and
process massive datasets and use these data to simulate large scale processes.
Simulations provide concrete and quantitative representations of a natural
phenomena relative to known input parameters, which can be perturbed to assess
the stability of data results. As a result, simulation experiments inspired by
observed data and domain knowledge are powerful tools to understand
how results may behave in real-world settings. They represent a best effort to
emulate complex processes, where the reliability of data results is not always
clear. Pairing such simulation studies with empirical evidence makes the DSLC more transparent for peers and users to review, aiding in the
objectivity of science.

\subsection{Stability at the modeling stage} 
Computational advances have fueled our ability to analyze the stability of data
results in practice.  At the modeling stage, stability measures how a data
result changes when the data and/or model are perturbed \cite{yu2013stability}. Stability extends the
concept of sampling variability in statistics, which is a measure of instability
relative to other data that could be generated from the same distribution.
Statistical uncertainty assessments implicitly assume stability in the form of a
distribution that generated the data. This assumption highlights the importance
of other data sets that could be observed under similar conditions (e.g. by
another person in the lab or another lab at another time). 

The concept of a model (``true'') distribution\footnote{We believe it is important to use the term ``model distribution'' instead of ``true distribution'' to avoid confusion over whether it is well-justified.} is a construct. When randomization is explicitly carried out, the model distribution can be viewed as a physical construct. Otherwise, it is a mental construct that must be justified
through domain knowledge, an understanding of the data generating process, and 
downstream utility. Statistical inference procedures use distributions to
draw conclusions about the real world. The relevance of such conclusions requires empirical support for the postulated model distribution,
especially when it is a mental construct.  In data science and statistical
problems, practitioners often do not make much of an attempt to justify this mental construct. At the same time, they take the
uncertainty conclusions very seriously. This flawed practice is likely related
to the high rate of false discoveries \cite{stark2018cargo,ioannidis2005most}.
It is a major impediment to progress of science and to data-driven
knowledge extraction in general.

While the stability principle encapsulates uncertainty quantification when the model distribution construct is well supported, it is intended to cover a much broader range of perturbations, such as problem formulation (e.g. different problem translations), pre-processing, EDA, randomized
algorithms, and choice of models/algorithms. Although seldom carried out in practice, evaluating stability across the entire DSLC is necessary to ensure that results are reliable and reproducible.  For example, the biologist studying gene regulation must choose both how to normalize raw data and what algorithm(s) she will use in her
analysis.  When there is no principled approach to make these decisions, the
knowledge data scientists can extract from analyses is limited to conclusions
that are stable across appropriate choices \cite{steegen2016increasing,
Abbasi2018deeptune, basu2018iterative}. This ensures that another scientist
studying the same data will reach similar conclusions, despite slight variation
in their independent choices.

\subsubsection{Formulating stability at the modeling stage} 
Stability at the modeling stage is defined with respect to a target of interest,
an appropriate perturbation to the data and/or
algorithm/model, and a stability metric to measure the change in target that
results from perturbation. We describe each of these in detail below.

\vspace{0.5em}
\noindent \textbf{Stability target:} The stability target
\begin{equation}
  \mathcal{T}(D, \lambda),
\end{equation}
corresponds to the data result or estimand of interest. It depends on input data $D$ and a
specific model/algorithm $\lambda$ used to analyze the data. For simplicity, we
will sometimes suppress the dependence on $D$ and $\lambda$ in our notation. As
an example, $\mathcal{T}$ can represent responses predicted by $h^{(\lambda)}$.  
Other examples of $\mathcal{T}$
include features selected by lasso with penalty parameter $\lambda$ or saliency
maps derived from a convolutional neural network (CNN) with architecture $\lambda$.

\vspace{0.5em}
\noindent\textbf{Data and model/algorithm perturbations:} To evaluate the stability of a
data result, we measure the change in target $\cal{T}$ that results from a
perturbation to the input data or learning algorithm. More precisely, we define
a collection of data perturbations $\mathbf{D}$ and model/algorithm
perturbations $\Lambda$ and compute the stability target distribution
\begin{equation}
  \{\mathcal{T} (D, \lambda) : D\in\mathbf{D}, \lambda \in \Lambda\}.
  \label{eq:stadist}
\end{equation}
For example, appropriate data perturbations include bootstrap sampling when
observations are approximately i.i.d., block bootstrap for weakly dependent time series, generative models that are supported
by domain knowledge (\sref{sec:data}), and probabilistic models that are justified from an
understanding of the data generating process or explicit randomization. When
different prediction functions are deemed equally appropriate based on domain
knowledge, each may represent an appropriate model perturbation (\sref{sec:model}).

It can be argued that the subjectivity surrounding appropriate
perturbations makes it difficult to evaluate results within the PCS framework.
Indeed, perturbation choices are both subjective human judgment calls and
critical considerations of PCS. The degree to which a data result can be trusted
depends on the justification for a perturbation. This is true if the
perturbation comes from a probabilistic model, as in traditional statistical
inference, or some broader set of perturbations, as in PCS. The goal of PCS is
to use and explicitly document perturbations that are best suited to assess
stability in complex, high-dimensional data rather than relying on probabilistic
models alone, which have little objective meaning when the model is not
justified. To ensure that results can be
evaluated, the case for an appropriate perturbation must be made in the
publication and in the PCS documentation (\sref{sec:doc}). These
transparent narratives allow readers to scrutinize and discuss perturbations to
determine which should be applied for a particular field and/or type of data,
encouraging objectivity.

\vspace{0.5em}
\noindent \textbf{Stability evaluation metric:} The stability evaluation metric
$s(\mathcal{T}; \mathbf{D}, \Lambda)$ summarizes the stability target
distribution in \eqref{eq:stadist}. For example, if $\mathcal{T}$ indicates
features selected by a model trained on data $D$, we may report the
proportion of times each feature is selected across data perturbations $D\in \mathbf{D}$. If $\mathcal{T}$ corresponds to saliency maps derived from different CNN architectures $\lambda\in\Lambda$, we may report each pixel's range of salience across $\Lambda$. When the stability evaluation metric combines targets across model/algorithm perturbations, it is important that these different targets are scaled appropriately to ensure comparability.

\vspace{0.5em}
A stability analysis that reveals the target
$\mathcal{T}$ is unstable (relative to a meaningful threshold for a particular
domain) may suggest an alternative analysis or target of interest.  This raises
issues of multiplicity and/or overfitting if the same data are used to evaluate new stability targets. Held-out test data offer one way to mitigate
these concerns. That is, training data can be used to identify a collection of
targets that are suitably stable. These targets can then be evaluated on the test data. More broadly, the process of refining analyses and
stability targets can be viewed as part of the iterative approach to data
analysis and knowledge generation described by \cite{box1976science}. Before
defining a new target or analysis, it may be necessary to collect new data to
help ensure reproducibility and external validity.

\subsubsection{Data perturbation}\label{sec:data}

The goal of data perturbation under the stability principle is to mimic a
process that could have been used to produce model input data but was not. This
includes human decisions, such as preprocessing and data cleaning, as well as
data generating mechanisms. When we focus on the change in target under possible
realizations of the data from a well-supported probabilistic model, we arrive at
well-justified sampling variability considerations in statistics. Hence data
perturbation under the stability principle includes, but is much broader than,
the concept of sampling variability.  It formally recognizes many other
important considerations in the DSLC beyond
sampling variability.  Furthermore, it provides a framework to assess trust in estimates of $\mathcal{T}$ when a probabilistic model is not
well-justified and hence sampling interpretations are not applicable.

Data perturbations can also be used to reduce variability in the estimated
target, which corresponds to a data result of interest. Random forests
incorporate subsampling data perturbations (of both the data units and
predictors) to produce predictions with better generalization error
\cite{breiman2001random}. Generative adversarial networks (GANs) use synthetic
adversarial examples to re-train deep neural networks and produce predictions
that are more robust to such adversarial data points
\cite{goodfellow2014generative}. Bayesian models based on conjugate priors
lead to marginal distributions that can be derived by adding observations to the
original data. Thus they can be viewed as a form of data perturbation that implicitly
introduces synthetic data through the prior. Empirically supported generative
models, including PDEs, can be used to explicitly introduce synthetic data.  As
with Bayesian priors, synthetic data perturbations from generative models can be
used to encourages stability of data results relative to prior knowledge, such
as mechanistic rules based on domain knowledge (for examples see
\cite{biegler2003large}).

\subsubsection{Algorithm or model perturbation}\label{sec:model} 
The goal of algorithm or model perturbation is to understand how alternative
analyses of the same data affect the target estimate. A classical example of
model perturbation is from robust statistics, where one searches for a robust
estimator of the mean of a location family by considering alternative models
with heavier tails than the Gaussian model.  Another example of model
perturbation is sensitivity analysis in Bayesian modeling
\cite{skene1986bayesian, box1980sampling}. Many of the model conditions used in
causal inference are in fact stability concepts that assume away confounding
factors by asserting that different conditional distributions are the same
\cite{peters2016causal, heinze2018invariant}.

Modern algorithms often have a random component, such as random projections or
random initial values in gradient descent and stochastic gradient descent. These
random components provide natural model perturbations that can be used to assess
the stability of $\mathcal{T}$. In addition to the random components of a single
algorithm, multiple models/algorithms can be used to evaluate stability of the
target. This is useful when there are many appropriate choices of
model/algorithm and no established criteria or established domain knowledge to
select among them. The stability principle calls for interpreting only the
targets of interest that are stable across these choices of algorithms or models
\cite{Abbasi2018deeptune}.

As with data perturbations, model perturbations can help reduce variability or
instability in the target. For instance, \cite{meinshausen2010stability} selects
lasso coefficients that are stable across different regularization parameters.
Dropout in neural networks is a form of algorithm perturbation that leverages
stability to improve generalizability \cite{srivastava2014dropout}. Our previous work
\cite{basu2018iterative} stabilizes random forests to interpret decision
rules in tree ensembles
\cite{basu2018iterative,kumbier2018refining}, which are perturbed using random
feature selection (model perturbation) and bootstrap (data perturbation).

\subsection{Dual roles of generative models in PCS}\label{sec:generative}

Generative models include both probabilistic models and partial differential
equations (PDEs) with initial or boundary conditions.
These models play dual roles in the PCS framework. On one hand, they can concisely summarize
past data and prior knowledge. On the other hand, they can be used to generate
synthetic observations that offer a form of data perturbation. 

When a generative model is used to summarize data, a common target of interest
is the model's parameters. Generative models with known parameters may be used
for prediction or to advance understanding through the mechanistic rules they
represent. Such models correspond to infinite data, though finite under
computational constraints. Generative models with unknown parameters can be used
to motivate surrogate loss functions through maximum likelihood and Bayesian
modeling methods. Mechanistic interpretations of such models should not be used
to draw scientific conclusions. They are simply useful starting points to
optimize algorithms that must be subjected to empirical validation.

Generative models that approximate the data generating process, a human judgment
call argued for in the PCS documentation, can be used as a form of data perturbation. Here synthetic data augment
the observed data and serve the purpose of domain-inspired regularization.  The
amount of synthetic data to combine with the observed data reflects our degree
of belief in the models, and is an interesting area for future exploration. Using synthetic data for domain inspired regularization
allows the same algorithmic and computing platforms to be applied to the
combined data.  This style of analysis is reminiscent of AdaBoost, which use the current data and model to modify the data used in the
next iteration without changing the base-learner \cite{freund1997decision}.

\subsection{Connections among the PCS principles}\label{sec:comp} 
Although we have discussed the three principles of PCS individually,
they share important connections. Computational considerations can limit the
predictive models/algorithms that are tractable, particularly for large,
high-dimensional datasets. These computability issues are often addressed in
practice through scalable optimization methods such as gradient descent (GD) or
stochastic gradient descent (SGD). Evaluating predictability on held-out data is
a form of stability analysis where the training/test sample split represents a
data perturbation. Other perturbations used to assess stability require multiple
runs of similar analyses. Parallel computation is well suited for these
perturbations.
\section{PCS inference through perturbation analysis}\label{sec:inference}

When data results are used to guide future decisions or actions, it is important
to assess the quality of the target estimate. For instance, suppose a model
predicts that an investment will generate a $10\%$ return over one year.
Intuitively, this prediction suggests that ``similar'' investments return $10\%$ on average. Whether or not a particular investment will realize a return close to $10\%$ depends on whether returns for ``similar'' investments ranged from $-20\%$ to $40\%$ or from $8\%$ to $12\%$. In other words, the variability of a prediction conveys important information about how much one should trust it. 

In traditional statistics, confidence measures describe the uncertainty of
an estimate due to sampling variability under a well-justified probabilistic
model.  \kkn{However, decisions made throughout the DSLC add another layer of uncertainty that may bias data results. This issue has been previously acknowledged in the modeling stage by \cite{coker2018theory}, who derive ``hacking intervals'' to assess the range of a summary statistic optimized over a possible set of data and algorithm perturbations.} In the PCS framework, we propose \textit{perturbation intervals}, or \textit{perturbation regions} in general, to quantify the stability of target estimates relative to different perturbations, including data cleaning and problem translations.
Perturbation intervals are conceptually similar to confidence intervals. The
primary difference is that they are explicitly connected to perturbations, justified in PCS documentation (\sref{sec:doc}) and evaluated by
independent reviewers and domain experts. 

As an example, perturbation intervals
for a target parameter from a single method based on bootstrap sampling specialize to traditional
confidence intervals based on the bootstrap. More broadly, perturbation
intervals quantify the variability of a target parameter value across the entire DSLC. For instance, a data scientist may consider multiple 
preprocessing, subsampling, and modeling strategies to predict investment
returns. The resulting perturbation intervals describe the range of
returns across worlds represented by each perturbation. Their reliability lies
squarely on whether the set of perturbations captures the full spectrum of
appropriate choices that could be made throughout the DSLC, which should be evaluated by domain experts and independent reviewers. This highlights the importance of perturbations that could plausibly generate the observed data, represent the range of uncertainty surrounding an analysis to the best degree possible, and are transparently documented for others to evaluate (\sref{sec:doc}).

As a starting point, we focus on a basic form of PCS inference that generalizes traditional statistical inference. Our approach to inference allows for a range of data and algorithm/model perturbations, making it flexible in its ability to represent uncertainty throughout the DSLC.

\subsection{PCS perturbation intervals}\label{sec:stabint} 
The reliability of perturbation intervals lies on the appropriateness of
each perturbation. Consequently, perturbation choices should be seriously
deliberated, clearly communicated, and evaluated by objective reviewers. Here we propose a framework for PCS inference based on a single problem translation and target estimand, leaving the case of multiple translations/estimands to future work.\footnote{The PCS perturbation intervals cover different problem translations through $\Lambda$ and are clearly extendable to
include perturbations in the pre-processing step through $\mathbf{D}$.}

\begin{enumerate}

  \item \textbf{Problem formulation:} Translate the domain question into a data
    science problem that specifies how the question will be addressed. Define a prediction target $y$, 	appropriate data
    $\mathbf{D}$ and/or model $\Lambda$ perturbations, prediction function(s)
    \{$h^{(\lambda)}:\lambda\in \Lambda\}$, training/test split, prediction
    evaluation metric $\ell$, stability metric $s$, and stability target
    $\mathcal{T}(D, \lambda)$. Document why these choices
    are appropriate in the context of the domain question.

  \item \textbf{Prediction screening:} For a threshold $\tau$, screen out models
    that do not fit the data (via prediction accuracy)
    \begin{equation}
      \Lambda^* = \{\lambda   \in \Lambda: \ell(h^{(\lambda)}, \mathbf{x}, y) < \tau\}.
      \label{eq:screen_pred}
    \end{equation}
    Examples of appropriate threshold include domain accepted baselines, the top
    $k$ performing models, or models whose accuracy is suitably similar to the
    most accurate model. If the goal of an analysis is prediction, testing data
    should be held-out until reporting the final prediction accuracy of a model in  step 4. In such a case, \eqref{eq:screen_pred} can be evaluated using a surrogate
    sample-splitting approach such as CV. If the goal of an analysis extends beyond prediction (e.g. to feature selection), \eqref{eq:screen_pred} may be evaluated on held-out test data.
    
  \item \textbf{Target value perturbation distributions:} For each of the
    survived models $\Lambda ^*$ from step 2, compute the stability
    target under each data perturbation $\mathbf{D}$. This results in a joint
    distribution of the target over data and model perturbations as in
    \eqref{eq:stadist}.  \kkn{For a collection of perturbations, requiring stability of $\mathcal{T}$ across all perturbations is more conservative in terms of type I error than requiring stability for any single perturbation. However, different domain questions require control over different types of error. How and when to combine results across perturbations is thus a human judgment call that should be transparently justified and documented.}
  
  \item \textbf{Perturbation result reporting:} Summarize the
    target value perturbation distribution using the stability metric $s$. \kkn{For instance, if $\mathcal{T}$ is one-dimensional we could summarize its perturbation distribution using the 10th and 90th percentiles or a visualization. If $\mathcal{T}$ is multi-dimensional, we could report a low dimensional projection of the perturbation distribution.} When perturbation results combine targets across models/algorithms, they may need to be rescaled to ensure comparability. When perturbation intervals are reported separately for model/algorithm perturbation, predictive accuracy evaluated in step 2 may be used as a measure of trust to rank each interval.

\end{enumerate}
At a high level, the PCS inference uses perturbation intervals to identify the stable part of accurate models. If perturbation results reveal instability among accurate
models, PCS inference can be used to interpret aspects that are shared (i.e.
stable) across these models. In this setting, PCS can be viewed as an implicit
application of Occam's razor. That is, it draws conclusions from the stable
portion of predictive models to simplify data results, making them more
reliable and easier to interpret. If perturbation intervals reveal that complex
models are both stable and accurate, PCS inference provides justification for the added complexity.

\subsection{PCS hypothesis testing}
Hypothesis testing from traditional statistics is commonly used in decision
making for science and business alike. The heart of Fisherian testing
\cite{fisher1992statistical} lies in calculating the p-value, which represents
the probability of an event more extreme than in the observed data under a null
hypothesis or distribution. Smaller p-values correspond to stronger evidence
against the null hypothesis or (ideally) the scientific theory embedded in the null
hypothesis. For example, we may want to determine whether a particular gene is
differentially expressed between breast cancer patients and a control group.
Given i.i.d. random samples from each population, we could address this question in the classical hypothesis testing framework using a t-test. The p-value
describes the probability of seeing a difference in means more extreme than
observed if the genes are not differentially expressed.

While hypothesis testing is valid philosophically, many of the assumptions
that it relies on are unrealistic in practice. For instance, unmeasured confounding
variables can bias estimates of causal effects. These issues are particularly
relevant in the social sciences, where randomized trials are difficult or
impossible to conduct. Resource constraints can limit how data are collected,
resulting in samples that do not reflect the population of interest, distorting the probabilistic interpretations of traditional statistical inference.
Moreover, hypothesis testing assumes empirical validity of probabilistic data
generating models.\footnote{Under conditions, Freedman
\cite{freedman1983nonstochastic} showed that some tests can be approximated by
permutation tests when data are not generated from a probabilistic model, but
these results are not broadly applicable.} When randomization is not carried
out explicitly, a particular null distribution must be justified from domain
knowledge of the data generating mechanism. Such issues are seldom taken
seriously in practice, resulting in settings where the null distribution is far
from the observed data. As a result, p-values as small as $10^{-5}$ or $10^{-8}$
are now common to report, despite the fact that there are rarely enough data to
reliably calculate these values, especially when multiple hypotheses (e.g.
thousands of genes) are evaluated.  When results are so far off on the tail of
the null distribution, there is no empirical evidence as to why the tail should
follow a particular parametric distribution. Moreover, hypothesis testing as
practiced today often relies on analytical approximations or Monte Carlo
methods, where issues arise for such small probability estimates. In fact,
there is a specialized area of importance sampling to deal with simulating small
probabilities \cite{rubino2009rare, bucklew2013introduction}, but these ideas have not been widely adopted in practice.

PCS hypothesis testing builds on perturbation intervals to address these practical issues and the cognitively misleading nature of small p-values.  It
uses the null hypothesis to define constrained perturbations that represent a
plausible data generating process, which in the best case corresponds to an existing scientific theory. This includes probabilistic models, when
they are well founded, as well as other data and/or algorithm perturbations. For
instance, generative models based on PDEs can be used to simulate data according
to established physical laws. Alternatively, a subset of data can be selected as controls (as an example see \cite{schuemie2018improving}). By allowing for a broad class of perturbations,
PCS hypothesis testing allows us to compare observed data with data that
respects some simple structure known to represent important characteristics of
the domain question. Of course, the appropriateness of a perturbation is a human
judgment call that should be clearly communicated in PCS documentation and
debated by researchers.  Much like scientists deliberate over appropriate
controls in an experiment, data scientists should debate the appropriate
perturbations in a PCS analysis.

\subsubsection{Formalizing PCS hypothesis testing} Formally, we consider
settings with observable input features $\mathbf{x}\in\mathcal{X}$, prediction
target $y\in\mathcal{Y}$, prediction functions
$\{h^{(\lambda)}: \lambda \in \Lambda\}$, and a null hypothesis that
qualitatively describes some aspect of the domain question. PCS hypothesis
testing translates the null hypothesis into a \textit{constrained perturbation}
and generates data
\begin{equation}
  D_0 = \{\mathbf{x}_0, y_0\}
\label{eq:nulldata}
\end{equation}
according to this perturbation.\footnote{A null hypothesis may correspond to
multiple data or model/algorithm perturbations. We focus on a single data
perturbation here for simplicity.} The particular choice of constrained
perturbation should be explicitly documented and justified by domain knowledge.
We use the constrained perturbation to construct and compare perturbation
intervals for both $D_0$ and $D$ and evaluate whether the observed data is
consistent with the hypothesis embedded in $D_0$.

\subsection{PCS inference in neuroscience and biology}
The work in \cite{elsayed2017structure} considers the null hypothesis that
population level structure in single neuron data is the expected byproduct of
primary features (e.g. correlations across time). This can be viewed as a form of PCS inference. The authors use a maximum
entropy approach, whose constraint is represented by the number of moments, to
generate data that share primary features with the observed data but are
otherwise random, and compare population level findings between the observed and
simulated data. In the accompanying PCS
\href{https://doi.org/10.5281/zenodo.3522419}{documentation}, we
consider the null hypothesis that genomic interactions appear with equal
frequency among different classes of genomic elements. We use a sample splitting
strategy which treats inactive elements (class-$0$ observations) as a baseline
to determine whether interactions appear with ``unusual'' frequency. Once again, these comparisons rely on human
judgment to determine when results are sufficiently different.
These choices depend on the domain context and how the problem has been
translated. They should be transparently communicated by the researcher in the
PCS documentation.

\subsection{PCS inference simulation studies in sparse linear models}
We tested PCS inference in an extensive set of data-inspired simulation experiments in the sparse linear model setting that has been widely studied by the statistics community over the past two decades (SI Appendix). In total, we considered 6 distinct generative models intended to reflect some of the issues that arise in practice. We compared our proposed PCS inference procedure with selective inference and asymptotic normality results using ROC curves. These provide a useful criterion to assess false positives and true positives, which are both important considerations in settings where resources dictate how many findings can be evaluated in follow-up analyses/experiments. Across all models, PCS inference compares favorably to both selective inference and asymptotic normality results (Fig. \ref{fig:simulation}). However, we note that the principal advantage of PCS inference is that it can be easily generalized to more complex settings faced by data scientists today as in the two examples described above.

\begin{figure}
\includegraphics[width=0.5\textwidth]{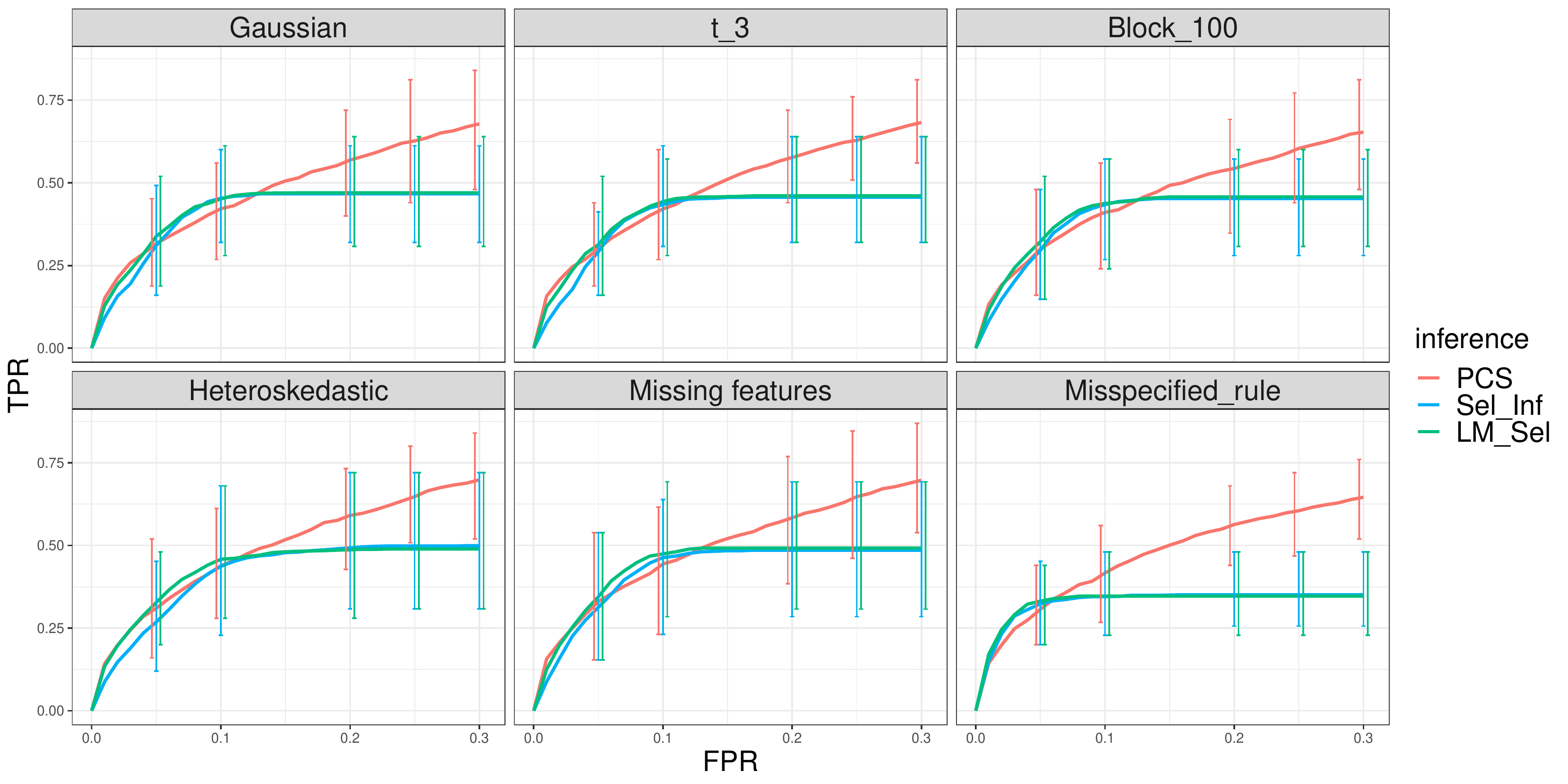}
\caption{ROC curves for feature selection in linear model setting with $n=250$ observations. Each plot corresponds to a different generative model.}
\label{fig:simulation}
\end{figure}

\section{PCS documentation}\label{sec:doc}

\begin{figure}
\centering
  \includegraphics[width=0.4\textwidth]{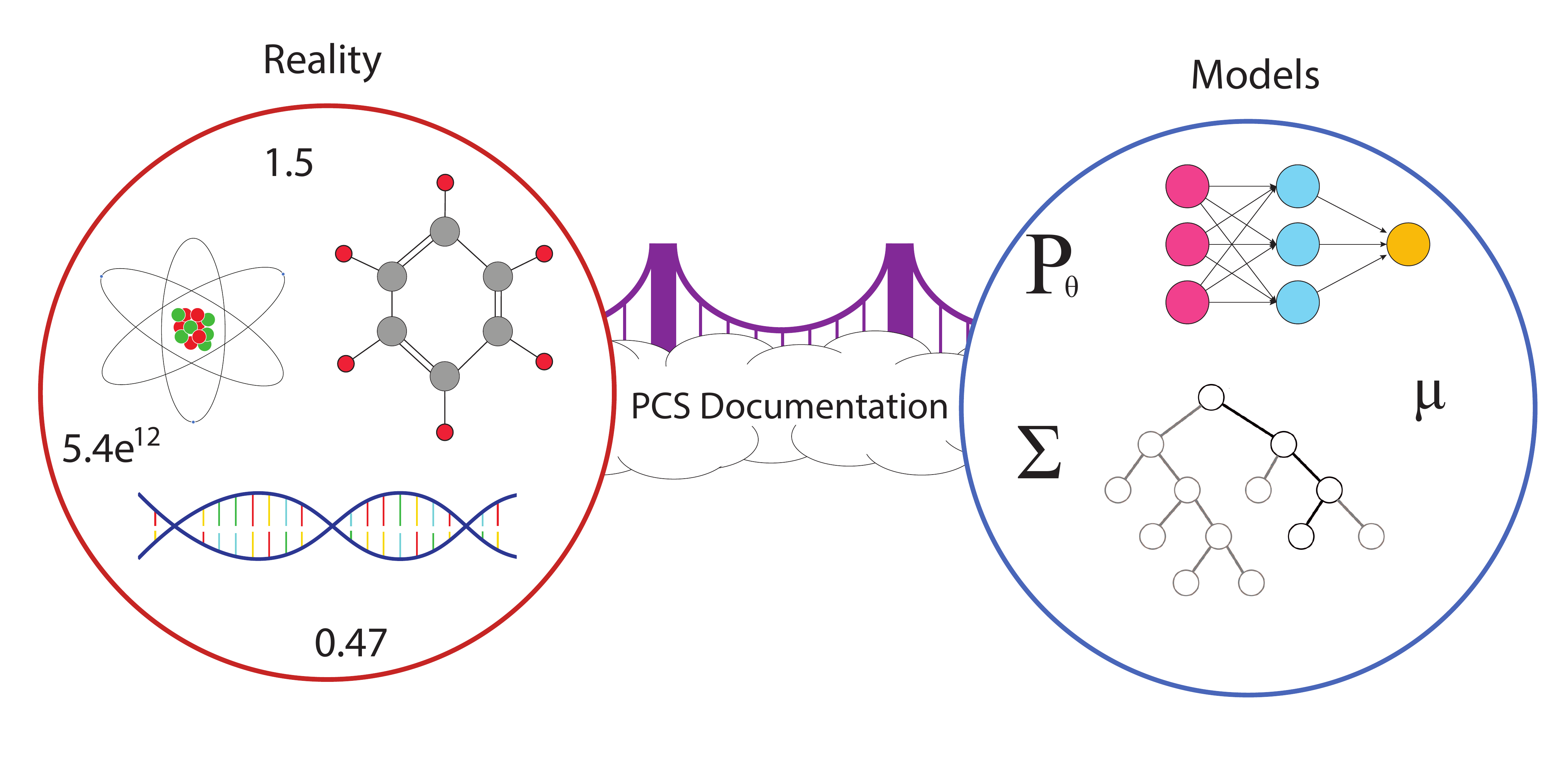}
  \caption{Assumptions made throughout the DSLC allow
  researchers to use models as an approximation of reality. Narratives provided
  in PCS documentation can help justify assumptions to connect these two
  worlds.}
  \label{fig:two_worlds}
\end{figure}

The PCS framework includes an accompanying R Markdown or Jupyter (iPython)
Notebook, which seamlessly integrates narratives, codes, and analyses. These
narratives are necessary to describe the domain problem and support assumptions
and choices made by the data scientist regarding computational platform, data
cleaning and preprocessing, data visualization, model/algorithm, prediction
metric, prediction evaluation, stability target, data and algorithm/model
perturbations, stability metric, and data conclusions in the context of the
domain problem. These narratives should be based on referenced prior knowledge
and an understanding of the data collection process, including design principles
or rationales. The narratives in the PCS documentation help
bridge or connect the two parallel universes of reality and models/algorithms
that exist in the mathematical world (Fig. \ref{fig:two_worlds}). In addition to
narratives justifying human judgment calls (possibly with data evidence), PCS
documentation should include all codes used to generate data results with links
to sources of data and metadata.

We propose the following steps in a notebook\footnote{This list is reminiscent
of the list in the ``data wisdom for data science" blog that one of the authors
wrote at http://www.odbms.org/2015/04/data-wisdom-for-data-science/}:
\vspace{-0.25em}
\begin{enumerate}
\item Domain problem formulation (narrative). Clearly state the real-world
  question and describe prior work related to this question. Indicate how this question can be answered in the context of a model or analysis.
\vspace{-0.5em}
\item Data collection and storage (narrative). Describe how the data were
  generated, including experimental design principles, and reasons why data is
  relevant to answer the domain question. Describe where data is stored and how it can be accessed by others.
\vspace{-0.5em}
\item Data cleaning and preprocessing (narrative, code, visualization). Describe
  steps taken to convert raw data into data used for analysis, and why 
  these preprocessing steps are justified. Ask whether more than one preprocessing
  methods should be used and examine their impacts on the final data results.
\vspace{-0.5em}
\item Exploratory data analysis (narrative, code, visualization). Describe 
    any preliminary analyses that influenced modeling decisions or conclusions along 
    with code and visualizations to support these decisions.
 \vspace{-0.5em}
\item Modeling and Post-hoc analysis (narrative, code, visualization). Carry out PCS inference in the context of the domain question.  Specify appropriate model and data perturbations. If necessary, specify null hypotheses and associated perturbations.
\vspace{-0.5em}
\item Interpretation of results (narrative and visualization). Translate the data results to draw conclusions and/or make recommendations in the context of domain problem.

\end{enumerate}

This documentation gives the reader as much information as
possible to make informed judgments regarding the evidence and process for
drawing a data conclusion in the DSLC. A case study of the PCS
framework in the genomics problem discussed earlier is documented on
\href{https://doi.org/10.5281/zenodo.3522419}{Zenodo}.

\section{PCS recommendation system for scientific hypothesis
generation}\label{sec:hypothesis}

In general, causality implies predictability and stability over many
experimental conditions; but not vice versa. The causal inference community has
long acknowledged connections between stability and estimates of causal effects.
For instance, many researchers have studied paradoxes surrounding associations
that lead to unstable estimates of causal effects \cite{fisher1923statistical,
cochran1938omission, bickel1975sex}. Estimates in the Neyman-Rubin potential
outcomes framework rely on a stable treatment across observational units
\cite{neyman1923applications, rubin1980randomization}.  Sensitivity analyses
test the stability of a causal effect relative to unmeasured confounding
\cite{cornfield1959smoking, ding2016sensitivity}.  Stability, particularly with respect to predictions across experimental interventions, has even been proposed as a criteria to establish certain causal relationships under the name ``invariance'' \cite{haavelmo1944probability, cartwright2003two, peters2016causal, scholkopf2012causal, pearl2009causality}.

PCS inference builds on these ideas, using stability and predictability to rank
target estimates for further studies, including follow-up experiments. In our
recent works on DeepTune \cite{Abbasi2018deeptune}, iterative random forests
(iRF) \cite{basu2018iterative}, and signed iterative random forests (siRF) \cite{kumbier2018refining}, we use PCS inference to make recommendations as inputs to downstream human decisions. For example, PCS inference suggested potential relationships between neurons in the visual cortex and visual stimuli as well as 3rd and 4th order interactions among biomolecules that
are candidates for regulating gene expression. Predictability and stability do
not replace physical experiments to prove or disprove causality.  However, we
hope computationally tractable analyses that demonstrate high predictability and
stability suggest hypotheses or intervention experiments that have higher yields
than otherwise.  This hope is supported by the fact that 80\% of the 2nd order
interactions identified by iRF \cite{basu2018iterative} had been verified in the literature through physical experiments.

\section{Conclusion}\label{sec:conc}

In this paper, we unified the principles of predictability, computability and stability (PCS) into a framework for veridical data science, comprised of both a workflow and documentation. The PCS framework aims to provide responsible, reliable, reproducible, and transparent results across the DSLC. It is a step towards systematic and unbiased inquiry in data science, similar to strong inference \cite{platt1964strong}. Prediction serves a reality check, evaluating how well a
model/algorithm captures the natural phenomena that generated the data.
Computability concerns with respect to algorithm efficiency determine the
tractability of the DSLC and point to the importance of data-inspired simulations in the design of useful algorithms.  Stability relative to data and model perturbations
was advocated in \cite{yu2013stability} as a minimum requirement for data
results' reproducibility and interpretability.

We made important conceptual progress on stability by extending it to the entire
DSLC, including problem formulation, data collection, data cleaning, and EDA. In
addition, we developed PCS inference to evaluate the variability of data results
with respect to a broad range of perturbations encountered in modern data
science. Specifically, we proposed PCS perturbation intervals to evaluate the reliability of data results and hypothesis testing to draw comparisons with simple structure in the data. We demonstrated that PCS inference performs favorably in a
feature selection problem through data-inspired sparse linear model simulation
studies and in a \href{https://doi.org/10.5281/zenodo.3522419}{genomics
case study}. To communicate the many human judgment calls in the DSLC, we
proposed PCS documentation, which integrates narratives justifying judgment
calls with reproducible codes and visualizations. This documentation makes
data-driven decisions as transparent as possible so that users of data results
can determine whether they are reliable.
 
In summary, we have offered a new conceptual and practical framework to guide
the DSLC, but many open problems remain. The basic PCS inference needs to be expanded into multi-translations of the same domain question and
vetted in practice well beyond the case studies in this paper and in our
previous works, especially by other researchers.  Additional case studies will
help unpack subjective human judgment calls in the context of specific domain
problems. The knowledge gained from these studies can be shared and critiqued
through transparent documentation. Based on feedback from practice, theoretical
studies of PCS procedures in the modeling stage are also called for to gain
further insights under stylized models after sufficient empirical vetting.
Finally, although there have been some theoretical studies on the connections
between the three principles (see \cite{hardt2015train, chen2018stability} and
references therein), much more work is necessary.

\section{Acknowledgements}
We would like to thank  Reza
Abbasi-Asl, Yuansi Chen, Ben Brown, Sumanta Basu, Adam Bloniarz,
Peter Bickel, Jas Sekhon, and S{\"o}reon K\"{u}nzel for
stimulating discussions on related topics in the past many years.  We also would
like to thank Richard Burk, Chris Holmes, Giles Hooker, Augie Kong, Avi Feller,
Peter B{\"u}hlmann, Terry Speed, Erwin Frise, Andrew Gelman, and Wei-cheng Kuo for their
helpful comments on earlier versions of the paper. We thank Tian Zheng, David Madigan, and the Yu group for discussions that led to the final title of this paper, which was suggested by Tian Zheng. Finally, we thank the reviewers for their extensive and helpful feedback. Partial supports are gratefully acknowledged from ARO grant W911NF1710005, ONR grant N00014-16-1-2664, NSF
grants DMS-1613002 and IIS 1741340, and the Center for Science of Information
(CSoI), a US NSF Science and Technology Center, under grant agreement
CCF-0939370. BY is a Chan Zuckerberg Biohub investigator.
\bibliographystyle{pnas-new}
\bibliography{pcs}

\begin{thebibliography}{10}

\bibitem{yu_kumbier_2018}
Yu B, Kumbier K (2018) Three principles of data science: predictability,
  computability, and stability (pcs):
  https://zenodo.org/record/1456199\#.xadgoodkjaj.

\bibitem{murdoch2019interpretable}
Murdoch WJ, Singh C, Kumbier K, Abbasi-Asl R, Yu B (2019) Definitions, methods,
  and applications in interpretable machine learning.
\newblock {\em Proceedings of the National Academy of Sciences}.

\bibitem{stark2018cargo}
Stark PB, Saltelli A (2018) Cargo-cult statistics and scientific crisis.
\newblock {\em Significance} 15(4):40--43.

\bibitem{ioannidis2005most}
Ioannidis JP (2005) Why most published research findings are false.
\newblock {\em PLoS medicine} 2(8):e124.

\bibitem{popperp1959logic}
Popperp KR (1959) {\em The Logic of Scientific Discovery}.
\newblock (University Press).

\bibitem{breiman2001statistical}
Breiman L, , et~al. (2001) Statistical modeling: The two cultures (with
  comments and a rejoinder by the author).
\newblock {\em Statistical science} 16(3):199--231.

\bibitem{stone1974cross}
Stone M (1974) Cross-validatory choice and assessment of statistical
  predictions.
\newblock {\em Journal of the royal statistical society. Series B
  (Methodological)} pp. 111--147.

\bibitem{allen1974relationship}
Allen DM (1974) The relationship between variable selection and data
  agumentation and a method for prediction.
\newblock {\em Technometrics} 16(1):125--127.

\bibitem{turing1937computable}
Turing AM (1937) On computable numbers, with an application to the
  entscheidungsproblem.
\newblock {\em Proceedings of the London mathematical society} 2(1):230--265.

\bibitem{hartmanis1965computational}
Hartmanis J, Stearns RE (1965) On the computational complexity of algorithms.
\newblock {\em Transactions of the American Mathematical Society} 117:285--306.

\bibitem{li2008introduction}
Li M, Vit{\'a}nyi P (2008) {\em An introduction to Kolmogorov complexity and
  its applications. Texts in Computer Science}.
\newblock (Springer, New York,) Vol.{}~9.

\bibitem{kolmogorov1963tables}
Kolmogorov AN (1963) On tables of random numbers.
\newblock {\em Sankhy{\=a}: The Indian Journal of Statistics, Series A} pp.
  369--376.

\bibitem{fisher1937design}
Fisher RA (1937) {\em The design of experiments}.
\newblock (Oliver And Boyd; Edinburgh; London).

\bibitem{donoho2009reproducible}
Donoho DL, Maleki A, Rahman IU, Shahram M, Stodden V (2009) Reproducible
  research in computational harmonic analysis.
\newblock {\em Computing in Science \& Engineering} 11(1).

\bibitem{stark2018before}
Stark P (2018) Before reproducibility must come preproducibility.
\newblock {\em Nature} 557(7707):613.

\bibitem{yu2013stability}
Yu B (2013) Stability.
\newblock {\em Bernoulli} 19(4):1484--1500.

\bibitem{manski2013public}
Manski CF (2013) {\em Public policy in an uncertain world: analysis and
  decisions}.
\newblock (Harvard University Press).

\bibitem{quenouille1949problems}
Quenouille MH, , et~al. (1949) Problems in plane sampling.
\newblock {\em The Annals of Mathematical Statistics} 20(3):355--375.

\bibitem{quenouille1956notes}
Quenouille MH (1956) Notes on bias in estimation.
\newblock {\em Biometrika} 43(3/4):353--360.

\bibitem{tukey1958bias}
Tukey J (1958) Bias and confidence in not quite large samples.
\newblock {\em Ann. Math. Statist.} 29:614.

\bibitem{efron1992bootstrap}
Efron B (1992) Bootstrap methods: another look at the jackknife in {\em
  Breakthroughs in statistics}.
\newblock (Springer), pp. 569--593.

\bibitem{bolstad2003comparison}
Bolstad BM, Irizarry RA, {\AA}strand M, Speed TP (2003) A comparison of
  normalization methods for high density oligonucleotide array data based on
  variance and bias.
\newblock {\em Bioinformatics} 19(2):185--193.

\bibitem{steegen2016increasing}
Steegen S, Tuerlinckx F, Gelman A, Vanpaemel W (2016) Increasing transparency
  through a multiverse analysis.
\newblock {\em Perspectives on Psychological Science} 11(5):702--712.

\bibitem{dawid1984present}
Dawid AP (1984) Present position and potential developments: Some personal
  views statistical theory the prequential approach.
\newblock {\em Journal of the Royal Statistical Society: Series A (General)}
  147(2):278--290.

\bibitem{freedman1991statistical}
Freedman DA (1991) Statistical models and shoe leather.
\newblock {\em Sociological methodology} pp. 291--313.

\bibitem{geisser1993predictive}
Geisser S (1993) {\em Predictive Inference}.
\newblock (CRC Press) Vol.{}~55.

\bibitem{tibshirani1996regression}
Tibshirani R (1996) Regression shrinkage and selection via the lasso.
\newblock {\em Journal of the Royal Statistical Society. Series B
  (Methodological)} pp. 267--288.

\bibitem{breiman2001random}
Breiman L (2001) Random forests.
\newblock {\em Machine learning} 45(1):5--32.

\bibitem{gneiting2007strictly}
Gneiting T, Raftery AE (2007) Strictly proper scoring rules, prediction, and
  estimation.
\newblock {\em Journal of the American Statistical Association}
  102(477):359--378.

\bibitem{wolpert1969positional}
Wolpert L (1969) Positional information and the spatial pattern of cellular
  differentiation.
\newblock {\em Journal of theoretical biology} 25(1):1--47.

\bibitem{fushiki2011estimation}
Fushiki T (2011) Estimation of prediction error by using k-fold
  cross-validation.
\newblock {\em Statistics and Computing} 21(2):137--146.

\bibitem{robbins1951stochastic}
Robbins H, Monro S (1951) A stochastic approximation method.
\newblock {\em The Annals of Mathematical Statistics} 22(3):400--407.

\bibitem{Abbasi2018deeptune}
Abbasi-Asl R, et~al. (2018) The deeptune framework for modeling and
  characterizing neurons in visual cortex area v4.
\newblock {\em bioRxiv} p. 465534.

\bibitem{basu2018iterative}
Basu S, Kumbier K, Brown JB, Yu B (2018) iterative random forests to discover
  predictive and stable high-order interactions.
\newblock {\em Proceedings of the National Academy of Sciences} p. 201711236.

\bibitem{box1976science}
Box GE (1976) Science and statistics.
\newblock {\em Journal of the American Statistical Association}
  71(356):791--799.

\bibitem{goodfellow2014generative}
Goodfellow I, et~al. (2014) Generative adversarial nets in {\em Advances in
  neural information processing systems}.
\newblock pp. 2672--2680.

\bibitem{biegler2003large}
Biegler LT, Ghattas O, Heinkenschloss M, van Bloemen~Waanders B (2003)
  Large-scale pde-constrained optimization: an introduction in {\em Large-Scale
  PDE-Constrained Optimization}.
\newblock (Springer), pp. 3--13.

\bibitem{skene1986bayesian}
Skene A, Shaw J, Lee T (1986) Bayesian modelling and sensitivity analysis.
\newblock {\em The Statistician} pp. 281--288.

\bibitem{box1980sampling}
Box GE (1980) Sampling and bayes' inference in scientific modelling and
  robustness.
\newblock {\em Journal of the Royal Statistical Society. Series A (General)}
  pp. 383--430.

\bibitem{peters2016causal}
Peters J, B{\"u}hlmann P, Meinshausen N (2016) Causal inference by using
  invariant prediction: identification and confidence intervals.
\newblock {\em Journal of the Royal Statistical Society: Series B (Statistical
  Methodology)} 78(5):947--1012.

\bibitem{heinze2018invariant}
Heinze-Deml C, Peters J, Meinshausen N (2018) Invariant causal prediction for
  nonlinear models.
\newblock {\em Journal of Causal Inference} 6(2).

\bibitem{meinshausen2010stability}
Meinshausen N, B{\"u}hlmann P (2010) Stability selection.
\newblock {\em Journal of the Royal Statistical Society: Series B (Statistical
  Methodology)} 72(4):417--473.

\bibitem{srivastava2014dropout}
Srivastava N, Hinton G, Krizhevsky A, Sutskever I, Salakhutdinov R (2014)
  Dropout: a simple way to prevent neural networks from overfitting.
\newblock {\em The Journal of Machine Learning Research} 15(1):1929--1958.

\bibitem{kumbier2018refining}
Kumbier K, Basu S, Brown JB, Celniker S, Yu B (2018) Refining interaction
  search through signed iterative random forests.
\newblock {\em arXiv preprint arXiv:1810.07287}.

\bibitem{freund1997decision}
Freund Y, Schapire RE (1997) A decision-theoretic generalization of on-line
  learning and an application to boosting.
\newblock {\em Journal of computer and system sciences} 55(1):119--139.

\bibitem{coker2018theory}
Coker B, Rudin C, King G (2018) A theory of statistical inference for ensuring
  the robustness of scientific results.
\newblock {\em arXiv preprint arXiv:1804.08646}.

\bibitem{fisher1992statistical}
Fisher RA (1992) Statistical methods for research workers in {\em Breakthroughs
  in statistics}.
\newblock (Springer), pp. 66--70.

\bibitem{freedman1983nonstochastic}
Freedman D, Lane D (1983) A nonstochastic interpretation of reported
  significance levels.
\newblock {\em Journal of Business \& Economic Statistics} 1(4):292--298.

\bibitem{rubino2009rare}
Rubino G, Tuffin B (2009) {\em Rare event simulation using Monte Carlo
  methods}.
\newblock (John Wiley \& Sons).

\bibitem{bucklew2013introduction}
Bucklew J (2013) {\em Introduction to rare event simulation}.
\newblock (Springer Science \& Business Media).

\bibitem{schuemie2018improving}
Schuemie MJ, Ryan PB, Hripcsak G, Madigan D, Suchard MA (2018) Improving
  reproducibility by using high-throughput observational studies with empirical
  calibration.
\newblock {\em Philosophical Transactions of the Royal Society A: Mathematical,
  Physical and Engineering Sciences} 376(2128):20170356.

\bibitem{elsayed2017structure}
Elsayed GF, Cunningham JP (2017) Structure in neural population recordings: an
  expected byproduct of simpler phenomena?
\newblock {\em Nature neuroscience} 20(9):1310.

\bibitem{fisher1923statistical}
Fisher RA (1923) Statistical tests of agreement between observation and
  hypothesis.
\newblock {\em Economica} (8):139--147.

\bibitem{cochran1938omission}
Cochran WG (1938) The omission or addition of an independent variate in
  multiple linear regression.
\newblock {\em Supplement to the Journal of the Royal Statistical Society}
  5(2):171--176.

\bibitem{bickel1975sex}
Bickel PJ, Hammel EA, O'Connell JW (1975) Sex bias in graduate admissions: Data
  from berkeley.
\newblock {\em Science} 187(4175):398--404.

\bibitem{neyman1923applications}
Neyman J (1923) Sur les applications de la th{\'e}orie des probabilit{\'e}s aux
  experiences agricoles: Essai des principes.
\newblock {\em Roczniki Nauk Rolniczych} 10:1--51.

\bibitem{rubin1980randomization}
Rubin DB (1980) Randomization analysis of experimental data: The fisher
  randomization test comment.
\newblock {\em Journal of the American Statistical Association}
  75(371):591--593.

\bibitem{cornfield1959smoking}
Cornfield J, et~al. (1959) Smoking and lung cancer: recent evidence and a
  discussion of some questions.
\newblock {\em Journal of the National Cancer institute} 22(1):173--203.

\bibitem{ding2016sensitivity}
Ding P, VanderWeele TJ (2016) Sensitivity analysis without assumptions.
\newblock {\em Epidemiology (Cambridge, Mass.)} 27(3):368.

\bibitem{haavelmo1944probability}
Haavelmo T (1944) The probability approach in econometrics.
\newblock {\em Econometrica: Journal of the Econometric Society} pp. iii--115.

\bibitem{cartwright2003two}
Cartwright N (2003) Two theorems on invariance and causality.
\newblock {\em Philosophy of Science} 70(1):203--224.

\bibitem{scholkopf2012causal}
Sch{\"o}lkopf B, et~al. (2012) On causal and anticausal learning.
\newblock {\em arXiv preprint arXiv:1206.6471}.

\bibitem{pearl2009causality}
Pearl J (2009) {\em Causality}.
\newblock (Cambridge university press).

\bibitem{platt1964strong}
Platt JR (1964) Strong inference.
\newblock {\em science} 146(3642):347--353.

\bibitem{hardt2015train}
Hardt M, Recht B, Singer Y (2015) Train faster, generalize better: Stability of
  stochastic gradient descent.
\newblock {\em arXiv preprint arXiv:1509.01240}.

\bibitem{chen2018stability}
Chen Y, Jin C, Yu B (2018) Stability and convergence trade-off of iterative
  optimization algorithms.
\newblock {\em arXiv preprint arXiv:1804.01619}.

\bibitem{macarthur2009developmental}
MacArthur S, et~al. (2009) Developmental roles of 21 drosophila transcription
  factors are determined by quantitative differences in binding to an
  overlapping set of thousands of genomic regions.
\newblock {\em Genome biology} 10(7):R80.

\bibitem{li2008transcription}
Li Xy, et~al. (2008) Transcription factors bind thousands of active and
  inactive regions in the drosophila blastoderm.
\newblock {\em PLoS biology} 6(2):e27.

\bibitem{li2014establishment}
Li XY, Harrison MM, Villalta JE, Kaplan T, Eisen MB (2014) Establishment of
  regions of genomic activity during the drosophila maternal to zygotic
  transition.
\newblock {\em Elife} 3:e03737.

\bibitem{taylor2015statistical}
Taylor J, Tibshirani RJ (2015) Statistical learning and selective inference.
\newblock {\em Proceedings of the National Academy of Sciences}
  112(25):7629--7634.

\end{thebibliography}

\section{Supporting Information: Simulation studies of PCS inference in the linear model setting}\label{sec:sims}
In this section, we consider the proposed PCS perturbation intervals through
data-inspired simulation studies. We focus on feature selection in sparse linear
models to demonstrate that PCS inference provides favorable results, in terms of
ROC analysis, in a setting that has been intensively investigated by the
statistics community in recent years. Despite its favorable performance in this
simple setting, we note that the principal advantage of PCS inference is its
generalizability to new situations faced by data scientists today. That is, PCS
can be applied to any algorithm or analysis where one can define appropriate
perturbations.  In the accompanying PCS case study, we demonstrate
the ease of applying PCS inference in the problem of selecting high-order,
rule-based interactions from a random forest in a high-throughput genomics 
problem (whose data the simulation studies below are based upon). 

To evaluate feature selection in the context of linear models, we considered
data for $35$ genomic assays measuring the binding enrichment of $23$ unique TFs
along $7809$ segments of the genome
\cite{macarthur2009developmental,li2008transcription,li2014establishment}.
That is, for an observation $\mathbf{x}_i = (x_{i1}, \dots, x_{ip})$,
$i=1\dots,n$, $x_{ij}$ measured the enrichment of the $j^{th}$ TF at the
$i^{th}$ segment of the genome. We augmented this data with $2^{nd}$ order
polynomial terms for all pairwise interactions (excluding quadratic terms $x_i ^
2$), resulting in a total of $p=35 + \binom{35}{2} = 630$ features. For a
complete description of the data, see the accompanying PCS
\href{https://zenodo.org/record/1456199#.XEjiE89KjRY}{documentation}.  We
standardized each feature and randomly selected $s=\lfloor \sqrt{p} \rfloor =
25$ active features to generate responses

\begin{equation}
y = \mathbf{x}^{T}\beta + \epsilon
\label{eq:lm}
\end{equation}
where $\mathbf{x}\in \mathbb{R} ^{7809 \times 630}$ denotes the normalized
matrix of features, $\beta_j=1$ for any active feature $j$ and $0$ otherwise,
and $\epsilon\in\mathbb{R}^n$ represents mean $0$ noise drawn from a variety of
distributions.  In total, we considered 6 distinct settings with 4 noise
distributions: i.i.d.  Gaussian, Students $t$ with $3$ degrees of freedom,
multivariate Gaussian with block covariance structure, Gaussian with variance
$\sigma^2_i\propto \|\mathbf{x}_i\|_2^2$ and two misspecified models: i.i.d.
Gaussian noise with 12 active features removed prior to fitting the model,
i.i.d. Gaussian noise with responses generated as

\begin{equation}
y = \sum_{S_j\in \mathcal{S}}\beta_{S_j} \prod_{k\in S_j} \mathds{1}(x_k > t_k) + \epsilon
\end{equation}
where $\mathcal{S}$ denotes a set of randomly sampled pairs of active features.

\subsection{Simple PCS perturbation intervals}
We evaluated selected features using the PCS perturbation intervals. 
Below we outline each step for constructing such
intervals in the context of linear model feature selection.

\begin{enumerate}
\item Our prediction target was the simulated responses $y$ and our stability
  target $\mathcal{T}\subseteq\{1,\dots,p\}$ the features selected by lasso when
    regressing $y$ on $\mathbf{x}$. To evaluate prediction accuracy, we
    randomly sampled $50\%$ of observations as a held-out test set. Our
    model/algorithm perturbatiosn were given by the default
    values of lasso penalty parameter in the R package \texttt{glmnet} 
    and $B=100$ bootstrap replicates respectively.

\item We formed a set of filtered models $\Lambda^*$ by taking $\lambda$
  corresponding to the $10$ most accurate models in terms of $\ell_2$ prediction
    error (Fig. \ref{fig:simulation}). Pre-specified prediction thresholds achieved qualitatively similar
    results and are reported in Fig. \ref{fig:simulation_nopred}. Since the goal of our analysis was feature selection, we evaluated
    prediction accuracy on the held-out test data. We repeated the steps below
    on each half of the data and averaged the final results.

\item For each $\lambda \in \Lambda^*$ and $b=1,\dots, 100$ we let
  $\mathcal{T}(\mathbf{x}^{(b)},\lambda)$ denote the features selected for bootstrap sample
    $b$ with penalty parameter $\lambda$.

\item The distribution of $\mathcal{T}$ across data and model perturbations can
  be summarized into a range of stability intervals. Since our goal was to
    compare PCS with classical statistical inference, which produces a single
    p-value for each feature, we computed a single stability score for each
    feature $j=1\dots,p$:
\end{enumerate}
\begin{equation*}
sta(j) = \frac{1}{B \cdot |\Lambda ^*|} \sum_{b=1}^{100} \sum_{\lambda \in
  \Lambda ^*} \mathds{1}(j\in \mathcal{T}(\mathbf{x}^{(b)},\lambda))
\end{equation*}
Intuitively, stability scores reflect our degree of belief that a given
feature is active in the model, with higher scores implying a higher degree of
certainty.  In practice, these scores could be used to rank features and
identify the most reliable collection for further consideration (e.g.
experimental validation). We note that the stability selection proposed in
\cite{meinshausen2010stability} is similar, but without the prediction error
screening.

\subsection{Results}
We compared the above PCS stability scores with asymptotic normality results
applied to features selected by lasso and selective inference
\cite{taylor2015statistical}.  We note that asymptotic normality and selective
inference both produce p-values for each feature, while PCS produces stability
scores.

\begin{figure*}
\begin{center}
\includegraphics[width=0.75\textwidth]{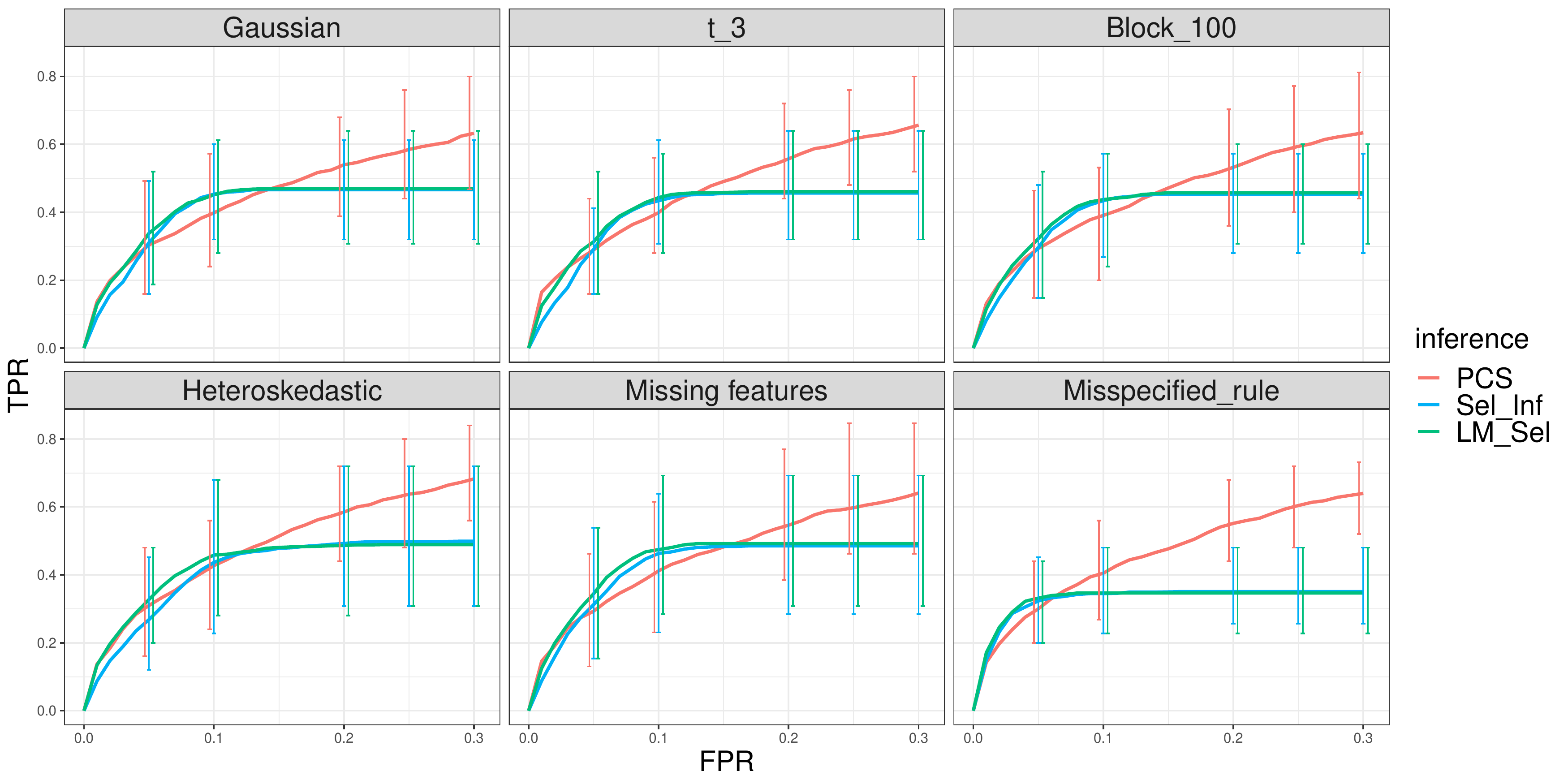}
\includegraphics[width=0.75\textwidth]{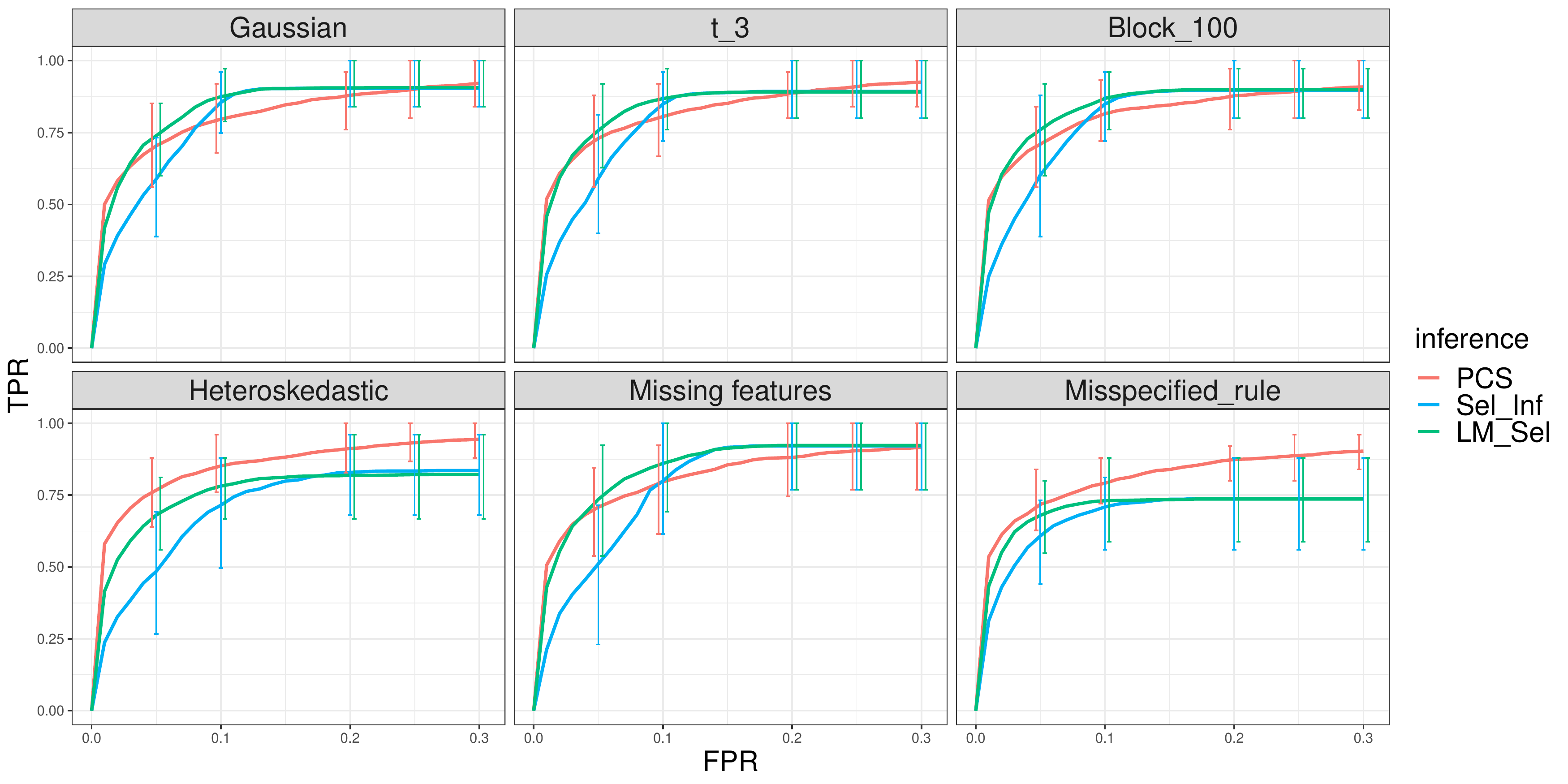}
\end{center}
\caption{ROC curves for feature selection in linear model setting, using pre-spcified threshold, with $n=250$ (top two rows) and $n=1000$ (bottom two rows) observations. Each plot corresponds to a different generative model. Prediction accuracy screening for PCS inference was conducted using a pre-specified threshold.}
\label{fig:simulation_nopred}
\end{figure*}

Figs. \ref{fig:simulation} and \ref{fig:simulation_nopred} show ROC curves for feature selection averaged across 100 replicates of the above experiments. The ROC curve is a useful evaluation
criterion to assess both false positive and true positive rates when
experimental resources dictate how many selected features can be evaluated in
further studies. In particular, ROC curves provide a balanced evaluation of each
method's ability to identify active features while limiting false discoveries.
Across all settings, PCS compares favorably to the other methods. The difference
is particularly pronounced in settings where other methods fail to recover a
large portion of active features ($n < p$, heteroskedastic, and misspecified
model). In such settings, stability analyses allow PCS to recover more active
features while still distinguishing them from inactive features. While its
performance in this setting is promising, the principal advantage of PCS is its
conceptual simplicity and generalizability. That is, the PCS perturbation
intervals described above can be applied in any setting where data or
model/algorithm perturbations can be defined, as illustrated in the genomics
case study in the accompanying PCS documentation. Traditional inference
procedures cannot typically handle multiple models easily.

\begin{figure*}
\begin{center}
\includegraphics[width=0.75\textwidth]{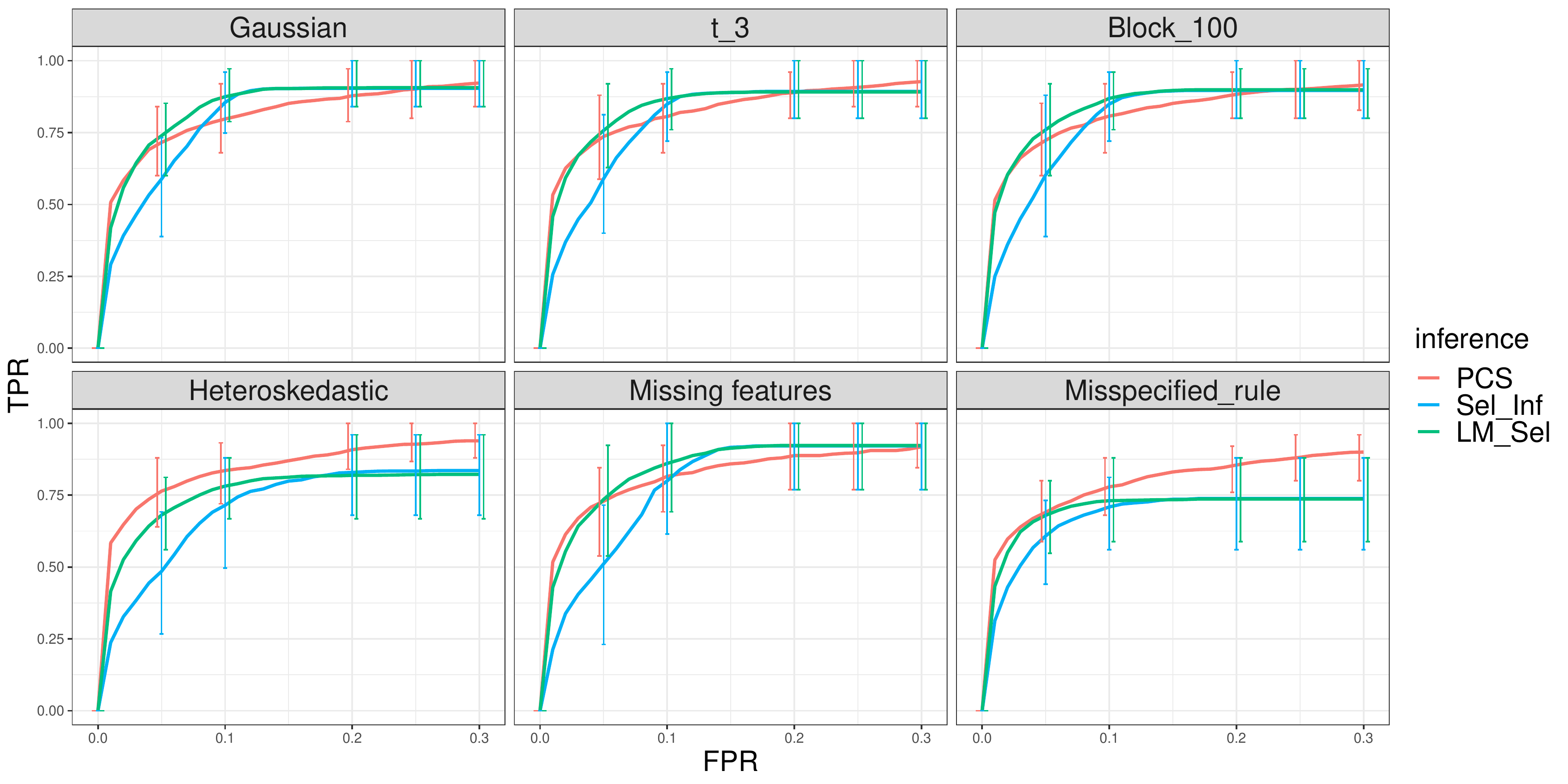}
\end{center}
\caption{ROC curves for feature selection in linear model setting, using 10 most accurate models, with $n=1000$ observations. Each plot corresponds to a different generative model. Prediction accuracy screening for PCS inference was conducted using a pre-specified threshold.}
\label{fig:simulation}
\end{figure*}

\end{document}